# Global Temporal Representation Based CNNs for Infrared Action Recognition

Yang Liu, Zhaoyang Lu, *Senior Member, IEEE*, Jing Li, *Member, IEEE*, Tao Yang, *Member, IEEE*, and Chao Yao

*Abstract*—Infrared human action recognition has many advantages, i.e., it is insensitive to illumination change, appearance variability, and shadows. Existing methods for infrared action recognition are either based on spatial or local temporal information, however, the global temporal information, which can better describe the movements of body parts across the whole video, is not considered. In this letter, we propose a novel global temporal representation named optical-flow stacked difference image (OFSDI) and extract robust and discriminative feature from the infrared action data by considering the local, global, and spatial temporal information together. Due to the small size of the infrared action dataset, we first apply convolutional neural networks on local, spatial, and global temporal stream respectively to obtain efficient convolutional feature maps from the raw data rather than train a classifier directly. Then these convolutional feature maps are aggregated into effective descriptors named three-stream trajectory-pooled deep-convolutional descriptors by trajectory-constrained pooling. Furthermore, we improve the robustness of these features by using the locality-constrained linear coding (LLC) method. With these features, a linear support vector machine (SVM) is adopted to classify the action data in our scheme. We conduct the experiments on infrared action recognition datasets InfAR and NTU RGB+D. The experimental results show that the proposed approach outperforms the representative state-of-the-art handcrafted features and deep learning features based methods for the infrared action recognition.

## I. Introduction

INFRARED action recognition has attracted increasing attention in recent years [1]–[5] because the infrared image can work well under dim light conditions due to its robustness to color and illumination changes, and cluttered background. Although convolutional neural networks (CNN) has achieved remarkable success for visible-light-based human action recognition [6]–[16], it cannot be directly applied to the infrared action recognition due to the lack of color and detailed appearance information in infrared images.

Recently, some deep learning based methods have been proposed for infrared action recognition. Gao *et al.* [3] adopted a two-stream CNN architecture to evaluate their newly built infrared action recognition dataset InfAR. This architecture consists of an optical-flow motion-history-image (OF-MHI) stream network and an OF stream network. Jiang *et al.* [4] developed a two-stream three-dimensional (3-D) CNN architecture to learn spatiotemporal features from the infrared videos. This architecture contains two separated recognition networks: Original infrared image net and OF net. They achieved better performance than that in [3] by introducing the 3-D CNN architecture and the discriminative code loss of the objective function. In these methods, the temporal features are local and generally consider only two (or a few) consecutive images in the video sequence. However, these local temporal information based methods tend to produce irrelevant information for action recognition such as scenery patterns and object patterns in the videos. Whereas taking the whole cycle of actions into consideration will lead to more attention on human actors in the videos, which can be considered as the global temporal information. Unlike the visible light action dataset, the available infrared action dataset is relatively small, which makes deep learning model difficult to be well trained [17], [18]. Thus, merely using deep learning framework as a feature extractor rather than a classifier can mitigate this problem.

Motivated by above observations, we build up a three-stream CNNs including local temporal stream, spatial-temporal stream, and global temporal stream for infrared action feature extraction. To our best knowledge, for action recognition in infrared videos, three-stream CNN architectures that take both the local and global temporal information into consideration have not been explored yet. In this letter, we propose a new architecture considering the global, local, and spatial-temporal information, simultaneously. The proposed architecture is based on the two-stream CNNs [8] used in trajectory-pooled deep convolutional descriptors (TDDs) [7]. Specifically, we use the temporal CNNs originated from the OF as local temporal stream, and replace the original spatial stream with one spatial-temporal stream and one global temporal stream via learning OF-MHI [19] CNNs features and the novel optical-flow stacked difference image (OFSDI) CNNs features, respectively. As CNNs architecture is very suitable to learn deep hidden features of complicated images [12], [20], we propose the OFSDI feature from the perspective of stacked difference of OF images to better

Manuscript received October 27, 2017; revised March 27, 2018; accepted April 4, 2018. Date of publication April 6, 2018; date of current version May 3, 2018. This work was supported in part by the National Natural Science Foundation of China under Grant 61502364 and Grant 61672429. The associate editor coordinating the review of this manuscript and approving it for publication was Dr. Xudong Jiang. *(Corresponding author: Jing Li.)*

Y. Liu, Z. Lu, and J. Li are with the School of Telecommunications Engineering, Xidian University, Xi'an 710071, China (e-mail: yliu_0@stu.xidian.edu.cn; zhylu@xidian.edu.cn; jinglixd@mail.xidian.edu.cn).

T. Yang is with the SAIIP, School of Computer Science, Northwestern Polytechnical University, Xi'an 710072, China (e-mail: tyang@nwpu.edu.cn).

C. Yao is with the School of Automation, Northwestern Polytechnical University, Xi'an 710072, China (e-mail: yaochao@nwpu.edu.cn).

Color versions of one or more of the figures in this letter are available online at http://ieeexplore.ieee.org.

Digital Object Identifier 10.1109/LSP.2018.2823910

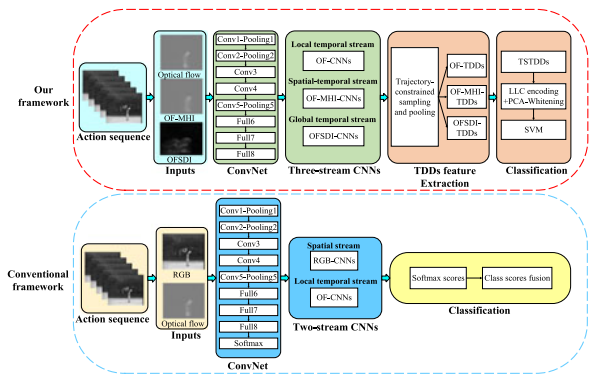

Fig. 1. Frameworks of our infrared action recognition method and the conventional method. Compared with the conventional method, our method is different in network input, CNNs structure, feature extraction, and classification strategy.

represent global temporal features of actions in deep. In this way, a three-stream CNNs is developed. Due to the limited size of infrared action dataset, the three-stream CNNs is then employed to learn an effective feature using trajectory-constrained sampling and pooling strategy [7] to share the benefits of both hand-crafted and deep-learned features. Finally, we choose the locality-constrained linear coding (LLC) [21] to aggregate the combined representation three-stream TDDs (TSTDDs) over the whole video into a global vector, then use linear support vector machine (SVM) [22] as the classifier to perform infrared action recognition on InfAR [2] and NTU RGB+D [23]. The contributions of this letter are threefold.

1) We propose a new formulation of global accumulated optical motion features named OFSDI to obtain the global temporal information of infrared action videos.
2) We propose a novel feature representation named TSTDDs for infrared action recognition incorporating the local temporal, spatial-temporal, and global temporal information.
3) To mitigate the problem that the existing deep learning model is difficult to be well trained due to the small size of the infrared action dataset, we use the three-stream CNNs framework as a feature extractor rather than a classifier.

The rest of this letter is organized as follows: Section II describes the proposed method. The experimental results are presented in Section III. Section IV concludes the letter with the remarks and future work.

## II. PROPOSED METHOD

Conventional deep learning methods for the infrared action recognition [4] usually predict actions via softmax scores from the softmax layers of network, shown in Fig. 1. However, the size of current infrared action dataset is significantly smaller than that of the visible light action dataset [7]. This will lead to a problem that the existing deep learning framework is difficult to be well trained, which has been validated by Nogueira *et al.* [17]. According to previous works [7], [18], [24], deep features can be extracted from the last layer before the classification layer and then used to train a linear classifier. Therefore, we use CNNs framework as a feature extractor rather than a classifier. The comparison of our method and conventional deep learning method for infrared action recognition can be seen in Fig. 1.

### A. Global Temporal Stream Based CNNs Configurations

To ensemble the global temporal information into the two-stream ConvNets[8] framework, we replace the original spatial CNNs with two independent CNNs, namely OF-MHI-CNNs and OFSDI-CNNs. The details of three-stream CNNs can be seen on the website https://xdyangliu.github.io/TSTDDs/.

The previous work [3] shows that the temporal information outperforms the appearance information in the infrared action recognition. However, to our best knowledge, there is no work that uses the whole cycle of an action for infrared action recognition. To address these problems, we propose a new formulation of global accumulated optical motion features named OFSDI, and use it as the input of the OFSDI-CNN networks. OFSDI aims to establish a global temporal template by fusing every local motion features over time. In this letter, we formulate it by absolute difference between the connecting OF images across the whole infrared action sequence, which can be defined as follows:

$$O(x, y, t) = \alpha O(x, y, t-1) + D(x, y, t) \quad (1)$$

where the parameter $\alpha$ is the update rate with $0 < \alpha < 1$, $O(x, y, t)$ and $O(x, y, t-1)$ are the OFSDI at time $t$ and $t-1$, respectively. And $D(x, y, t)$ is the image difference between OF images at time $t$ and $t-1$, it is defined as

$$D(x, y, t) = |F(x, y, t) - F(x, y, t-1)| \quad (2)$$

where $t \in [2, L]$, $L$ denotes the length of the video and $F(x, y, t)$ denotes the brightness of the pixel point $(x, y)$ in an OF image at time $t$.

Given (1) and (2), the OFSDI of a video frame at time $t$ can be represented in an accumulation form as follows:

$$O(t) = \alpha^t O(0) + \sum_{i=1}^{t} \alpha^{(t-i)} D(i) \quad (3)$$

where $O(0)$ is the initial image of the OFSDI and $\alpha^t (t \in \{1, \ldots, L\})$ controls its contribution to the formulation of OFSDI. In (3), the OFSDI is given by the accumulation of OF difference image across the whole video duration and the contribution of each OF difference image at time $t$ is controlled by $\alpha^{(t-i)}$. Intuitively, the recent OF difference images (temporal information) contributes much more to the OFSDI than the OF difference images at initial time. Thus, the OFSDI is global temporal and records the long-term motion information of an action. To match the OFSDI-CNN, the pixel value of $O(x, y, t)$ is transformed into $[0, 255]$ and replicates each OFSDI image to produce an image with three same channels in our method.

To better represent the actions in infrared videos, the spatial-temporal network OF-MHI-CNN and the local temporal network OF-CNN, are also considered. For OF-MHI-CNN network, we use OF-MHI image [19] as an input. For OF-CNN network, the input is 3-D volume of stacking OF images [25] computed using each consecutive pair of frames. Therefore, it is local temporal without considering the whole action procedure.

With the three-stream CNNs, given a video V, the aforementioned CNNs can generate three kinds of convolutional feature maps, which represent the video complementarily. The three-stream CNNs feature maps can be denoted as follows:

$$\mathbb{C}(V) = \{C_1^{lt}, \ldots, C_M^{lt}, C_1^{st}, \ldots, C_M^{st}, C_1^{gt}, \ldots, C_M^{gt}\} \quad (4)$$

where $C_m^{lt} \in \mathbb{R}^{H_m \times W_m \times L \times N_m}$, $C_m^{st} \in \mathbb{R}^{H_m \times W_m \times L \times N_m}$, and $C_m^{gt} \in \mathbb{R}^{H_m \times W_m \times L \times N_m}$ denote the $m$th feature maps of the local, spatial, and global temporal CNNs, individually. $H_m$, $W_m$, and $L$ denote the height, the width, and the duration of the video, $N_m$ is the number of channels, and $M$ is the number of layers of extracting TSTDDs. Specifically, we extract the feature maps from conv4 and conv5 for spatial-temporal network and global temporal network, and conv3 and conv4 layers for the local temporal network.

### B. Three-Stream Trajectory-Pooled Deep-Convolutional Descriptors

To obtain robust and discriminative video feature representation and incorporate video temporal characteristics into deep architectures, we will describe the process of extracting TSTDDs from a set of improved trajectories $\mathbb{T}(V)$ and convolutional feature maps $\mathbb{C}(V)$ for a given video $V$.

We first extract a set of improved trajectories [26] for each video $V$

$$\mathbb{T}(V) = \{T_1, T_2, \ldots, T_K\} \quad (5)$$

where $K$ is the number of trajectories, and $T_K$ denotes the $k$th trajectory in the original spatial scale

$$T_k = \{(x_1^k, y_1^k, z_1^k), (x_2^k, y_2^k, z_2^k), \ldots, (x_P^k, y_P^k, z_P^k)\} \quad (6)$$

where $(x_p^k, y_p^k, z_p^k)$ is the pixel position of the $p$th point in trajectory $T_k$, and $P$ is the length of the trajectory ($P = 15$).

To reduce the influence of illumination changes, we normalize the three-stream CNNs map $\mathbb{C}(V)$ by both spatiotemporal normalization and channel normalization methods following the work [7].

For spatiotemporal normalization, we normalize the convolutional feature map $C \in \mathbb{R}^{H \times W \times L \times N}$ of (4) as follows:

$$\widetilde{C}_{st}(x, y, z, n) = C(x, y, z, n) / \max V_{st}^n \quad (7)$$

where $\max V_{st}^n = \max_{x,y,z} C(x, y, z, n)$ is the maximum value of the $n$th feature maps over the whole video spatiotemporal extent.

For channel normalization, we normalize the convolutional feature map $C \in \mathbb{R}^{H \times W \times L \times N}$ of (4) as follows:

$$\widetilde{C}_{ch}(x, y, z, n) = C(x, y, z, n) / \max V_{ch}^{x,y,z} \quad (8)$$

where $\max V_{ch}^{x,y,z} = \max_n C(x, y, z, n)$ is the maximum value of different feature channels at pixel position $(x, y, z)$.

After feature normalization, we extract TSTDDs based on trajectories and normalized convolutional feature maps by using trajectory pooling. Specifically, given a trajectory $T_k$ and a normalized feature map $\widetilde{C}_m^a$, which is the $m$th layer feature map after either spatiotemporal normalization or channel normalization from the local temporal net, spatial-temporal net, or global temporal net ($a \in \{lt, st, gt\}$). We conduct sum-pooling of the normalized feature maps over the 3-D volume centered at the trajectory as follows:

$$D(T_k, \widetilde{C}_m^a) = \sum_{p=1}^{P} \widetilde{C}_m^a(\overline{(r_m \times x_p^k)}, \overline{(r_m \times y_p^k)}, z_p^k) \quad (9)$$

where $(x_p^k, y_p^k, z_p^k)$ is the $p$th point position of trajectory $T_k$, $r_m$ is the $m$th layer map size ratio $\overline{(\cdot)}$ is the rounding operation. $D(T_k, \widetilde{C}_m^a)$ is called the TDDs. Due to the complementary property of these two normalization methods validated by [7], we use the combined representation obtained from these two normalization methods for TDDs.

By the above mentioned approach, we can obtain the OF-TDDs, OF-MHI-TDDs, and OFSDI-TDDs, respectively from the OF-CNNs, the OF-MHI-CNNs, and the OFSDI-CNNs. To keep the dimensionality of TSTDDs in a reasonable size, we fix the dimension 256 for each of them reduced by principal component analysis (PCA). Finally, the TSTDDs is obtained by the concatenation of them with dimension Dim = 768. Then, we adopt the LLC [21] scheme to represent the TSTDDs by five local bases, and the codebook size is set to be 4000 for all the training-testing partitions. To reduce the complexity when constructing the codebook, only 200 TSTDDs are randomly selected from each video sequence. After LLC encoding, we utilize the PCA to preprocess encoded TSTDDs features while guaranteeing the contribution rate to be higher than 99%. With the TSTDDs, we use a linear SVM [22] with linear kernel to discriminate the infrared action videos.

## III. EXPERIMENTAL RESULTS

### A. Dataset

The proposed method is evaluated on the first publicly available infrared action recognition dataset InfAR [2] and a large scale human action recognition dataset NTU RGB+D [23]. The InfAR dataset consists of 600 video sequences captured by infrared thermal imaging cameras. The action classes include fight, handclapping, handshake, hug, jog, jump, punch, push, skip, walk, wave1 (one hand wave), and wave2 (two hands wave). Each action class has 50 video clips and there are 600 video samples in total. NTU RGB+D dataset consists of 56 880 action samples captured by three Microsoft Kinect v.2 cameras concurrently. Due its huge size, we select ten infrared action classes A001–A010 from Setup 9 to evaluate our method. These action classes include drink water, eat meal/snack, brushing teeth, brushing hair, drop, pickup, throw, sitting down, standing up (from sitting position), and clapping. Each action class consists of 42 video samples and there are 420 video samples in total. Example images of these two datasets can be seen on the website https://xdyangliu.github.io/TSTDDs/.

### B. Experimental Settings

In our experiments, both InfAR and NTU RGB+D datasets are randomly split into training and testing sets. The experiments with the same setting are repeated for five times in each evaluation. For InfAR dataset, 30 samples in each class are randomly selected as training samples. While for NTU RGB+D dataset, half of the samples in each class are randomly selected as training samples. And the others are used for testing. For local temporal stream, we use the publicly available "temporal" network from [7], which has been pretrained on the UCF101 dataset [27]. For spatial-temporal and global temporal streams, the spatial network provided by [7] is used. And the parameters $\alpha$ and $O$ of (1) are empirically set as 0.96 and the first image of the original OF, respectively. For domain adaptation, the last fully connected layers of all the networks are fine-tuned on InfAR dataset or NTU RGB+D dataset via stochastic gradient descent. The initial learning rate is set as $10^{-3}$, and decrease to $10^{-4}$ after 3 K iterations. It is then reduced to $10^{-5}$ and $10^{-6}$ after 6 K and 9 K iterations, respectively. And the training is stopped at 10 K iterations. The minibatch size are 32 and 21 for InfAR dataset and NTU RGB+D dataset, respectively. The dropout ratio is chosen as 0.5 for both fc6 and fc7 layers. We employ Caffe toolbox for CNNs learning [28] and LibSVM toolbox [22] for SVMs implementation. The parameter $C$ of linear SVM is set as 30 by fivefold cross validation. We release all the data and codes on the website https://xdyangliu.github.io/TSTDDs/.

TABLE I
RECOGNITION RESULTS (%) OF EACH STREAM WITH DIFFERENT INPUT AND THEIR COMBINATIONS ON INFAR AND NTU RGB+D DATASETS

| Method | InfAR | NTU RGB+D |
|---|---|---|
| Spatial TDDs | 39.00 | 51.43 |
| OF-TDDs | 75.83 | 55.81 |
| OF-MHI-TDDs | 67.67 | 48.19 |
| OFSDI-TDDs | 62.58 | 35.24 |
| (Spatial+OF)-TDDs | 63.58 | 63.14 |
| (OF+OF-MHI)-TDDs | 75.58 | 65.05 |
| **TSTDDs** | **79.25** | **66.29** |

TABLE II
RECOGNITION RESULTS (%) OF DIFFERENT METHODS ON INFAR DATASET

| Feature type | Method | Average accuracy (%) |
|---|---|---|
| Hand-crafted | HOF [3] | 68.58 |
|  | Dense-Traj [29] | 68.66 |
|  | iDTs [26] | 71.83 |
| Deep learning | TDDs [7] | 63.58 |
|  | Two-stream CNNs [8] | 76.66 |
|  | 3D CNNs [4] | 77.50 |
|  | **TSTDDs** | **79.25** |

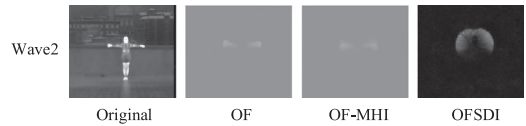

Fig. 2. Comparison of the original infrared, OF, OF-MHI and our proposed OFSDI images on InfAR dataset.

### C. Evaluations on InfAR and NTU RGB+D Datasets

Initially, we evaluate the performance of each stream of the three-stream CNNs, spatial stream with raw infrared images as input, some of their combinations, and our proposed TSTDDs on InfAR dataset, shown in Table I. Then, the performance of TSTDDs is compared with that of representative state-of-the-art handcrafted features based methods improved dense trajectories (iDTs) [26], Dense-Traj [29], HOF [3] and deep learning features based methods TDDs [7], Two-stream CNNs [8], and 3-D CNNs [4]. All the comparison results are summarized in Table II. For comparison purpose, all methods are implemented with the same experimental settings.

From Tables I and II, we can see that the performance of Spatial TDDs is the lowest (39.00%) and (Spatial+OF)-TDDs only achieves the accuracy of 63.58%, which is even much worse than the handcrafted-based methods, such as HOF, Dense-Traj, and iDTs. This demonstrates that the appearance information in original infrared images does not help too much for the action recognition. The performance of Dense-Traj and HOF are almost the same even though the HOF is one of the components of the Desnse-Traj, which indicates that the motion (temporal) information is important in the infrared action recognition, which can also be intuitively observed in Fig. 2. We can find that the unrelated background clutter can be removed in OF images, OF-MHI images, and OFSDI images. In addition, the proposed OFSDI is global temporal as it can capture the whole process of motion (long-time motion information), which is complementary to the local temporal information provided by the OF and the spatial-temporal information provided by the OF-MHI. The performance of three independent stream OF-TDDs, OF-MHI-TDDs, and OFSDI-TDDs are 75.83%, 67.67%, and 62.58%, respectively. The combination of them, TSTDDs, achieves as high as 79.25%, and outperforms other representative state-of-the-art methods, which demonstrates the effectiveness of the TSTDDs, and also indicates that the three stream CNNs are complementary. To be noticed, the performance of (OF+OF-MHI)-TDDs is 75.58% while TSTDDs is 79.25%. This illustrates that adding OFSDI module indeed improve the overall performance.

Due to the lack of the related work specified for infrared modality on NTU RGB+D dataset, we do not compare our method with other methods. Instead, we evaluate the performance of independent stream with different inputs, their combinations, and the TSTDDs. From Table I, we can see that the performance of OFSDI-TDDs is the lowest. This is because videos on NTU RGB+D dataset contains large view variations [23] and the global temporal feature captured by OFSDI is not view-invariant, which leads to performance degradation. Nevertheless, the combinations of the three independent streams, TSTDDs, achieves the best performance 66.29%. This demonstrates the effectiveness of the TSTDDs, and also indicates that the three stream CNNs are complementary. To be noticed, the performance of (OF+OF-MHI)-TDDs is 65.05% while TSTDDs is 66.29%. This illustrates that the global temporal information from OFSDI indeed improves the overall performance.

We execute our codes on Intel (R) CoreTM i7 system with 32-GB RAM and NVIDIA GTX960 GPU. For TSTDDs feature extraction, the average time for each video on InfAR and NTU RGB+D datasets are 24.50 and 16.29 s, respectively. For testing each video, the average time is 5.71 s, which includes 5.70 s for LLC encoding and PCA Whitening, and 0.01 s for the model to predict the class.

### IV. CONCLUSION

This letter proposes a novel global temporal representation based three-stream CNNs framework incorporating the local temporal, spatial-temporal, and global temporal information to learn a discriminative infrared video representation called TST-DDs for the infrared action recognition. The proposed method outperforms the representative state-of-the-art methods for the infrared action recognition on InfAR and NTU RGB+D datasets. Our future work is to extend the current approach to handle the problem of large view variations on more challenge datasets, and explore the transfer learning techniques that can learn view-invariant features to enhance the performance of the infrared action recognition.

## References

[1] Y. Zhu and G. Guo, "A study on visible to infrared action recognition," *IEEE Signal Process. Lett.*, vol. 20, no. 9, pp. 897–900, Sep. 2013.

[2] C. Gao, Y. Du, J. Liu, L. Yang, and D. Meng, "A new dataset and evaluation for infrared action recognition," in *Proc. CCF Chin. Conf. Comput. Vis. 2015*, pp. 302–312.

[3] C. Gao *et al.*, "Infar dataset: Infrared action recognition at different times," *Neurocomputing*, vol. 212, pp. 36–47, 2016.

[4] Z. Jiang, V. Rozgic, and S. Adali, "Learning spatiotemporal features for infrared action recognition with 3-D convolutional neural networks," in *Proc. Comput. Vis. Pattern Recognit. Workshops*, Jul. 2017, pp. 309–317.

[5] J. Li, C. Li, T. Yang, and Z. Lu, "A novel visual-vocabulary-translator-based cross-domain image matching," *IEEE Access*, vol. 5, pp. 23 190–23 203, 2017.

[6] S. Ji, W. Xu, M. Yang, and K. Yu, "3-D convolutional neural networks for human action recognition," *IEEE Trans. Pattern Anal. Mach. Intell.*, vol. 35, no. 1, pp. 221–231, Jan. 2013.

[7] L. Wang, Y. Qiao, and X. Tang, "Action recognition with trajectory-pooled deep-convolutional descriptors," in *Proc. Comput. Vis. Pattern Recognit.*, 2015, pp. 4305–4314.

[8] K. Simonyan and A. Zisserman, "Two-stream convolutional networks for action recognition in videos," in *Proc. Adv. Neural Inf. Process. Syst.*, 2014, pp. 568–576.

[9] H. Bilen, B. Fernando, E. Gavves, A. Vedaldi, and S. Gould, "Dynamic image networks for action recognition," in *Proc. Comput. Vis. Pattern Recognit.*, 2016, pp. 3034–3042.

[10] C. Feichtenhofer, A. Pinz, and A. Zisserman, "Convolutional two-stream network fusion for video action recognition," in *Proc. Comput. Vis. Pattern Recognit.*, 2016, pp. 1933–1941.

[11] G. Varol, I. Laptev, and C. Schmid, "Long-term temporal convolutions for action recognition," *IEEE Trans. Pattern Anal. Mach. Intell.*, to be published, doi: 10.1109/TPAMI.2017.2712608.

[12] L. Wang, L. Ge, R. Li, and Y. Fang, "Three-stream cnns for action recognition," *Pattern Recognit. Lett.*, vol. 92, pp. 33–40, 2017.

[13] S. Zhao, Y. Liu, Y. Han, R. Hong, Q. Hu, and Q. Tian, "Pooling the convolutional layers in deep convnets for video action recognition," *IEEE Trans. Circuits Syst. Video Technol.*, to be published, doi: 10.1109/TCSVT.2017.2682196.

[14] X. Wang, L. Gao, J. Song, and H. T. Shen, "Beyond frame-level CNN: Saliency-aware 3-D CNN with LSTM for video action recognition," *IEEE Signal Process. Lett.*, vol. 24, no. 4, pp. 510–514, Apr. 2017.

[15] C. Li, Y. Hou, P. Wang, and W. Li, "Joint distance maps based action recognition with convolutional neural networks," *IEEE Signal Process. Lett.*, vol. 24, no. 5, pp. 624–628, May 2017.

[16] Q. Chen and Y. Zhang, "Sequential segment networks for action recognition," *IEEE Signal Process. Lett.*, vol. 24, no. 5, pp. 712–716, May 2017.

[17] K. Nogueira, O. A. B. Penatti, and J. A. dos Santos, "Toward better exploiting convolutional neural networks for remote sensing scene classification," *Pattern Recognit.*, vol. 61, pp. 539–556, 2017.

[18] Y. Liu, Z. Lu, J. Li, C. Yao, and Y. Deng, "Transferable feature representation for visible-to-infrared cross-dataset human action recognition," *Complexity*, vol. 2018, pp. 1–20, 2018.

[19] D. Tsai, W. Chiu, and M. Lee, "Optical flow-motion history image (OF-MHI) for action recognition," *Signal, Image Video Process.*, vol. 9, no. 8, pp. 1897–1906, 2015.

[20] Q. Ke, M. Bennamoun, S. An, F. Sohel, and F. Boussaid, "Leveraging structural context models and ranking score fusion for human interaction prediction," *IEEE Trans. Multimedia*, to be published, doi: 10.1109/TMM.2017.2778559.

[21] J. Wang, J. Yang, K. Yu, F. Lv, T. S. Huang, and Y. Gong, "Locality-constrained linear coding for image classification," in *Proc. Comput. Vis. Pattern Recognit.*, 2010, pp. 3360–3367.

[22] C. Chang and C. Lin, "LIBSVM: A library for support vector machines," *ACM Trans. Intell. Syst. Technol.*, vol. 2, no. 3, pp. 27:1–27:27, 2011.

[23] A. Shahroudy, J. Liu, T. Ng, and G. Wang, "NTU RGB+D: A large scale dataset for 3-D human activity analysis," in *Proc. Comput. Vis. Pattern Recognit. Workshops*, 2016, pp. 1010–1019.

[24] A. S. Razavian, H. Azizpour, J. Sullivan, and S. Carlsson, "CNN features off-the-shelf: An astounding baseline for recognition," in *Proc. Comput. Vis. Pattern Recognit. Workshops*, 2014, pp. 512–519.

[25] C. Zach, T. Pock, and H. Bischof, "A duality based approach for realtime tv-$L^1$ optical flow," in *Proc. Pattern Recognit.*, 2007, pp. 214–223.

[26] H. Wang and C. Schmid, "Action recognition with improved trajectories," in *Proc. Int. Conf. Comput. Vis.*, 2013, pp. 3551–3558.

[27] K. Soomro, A. R. Zamir, and M. Shah, "UCF101: A dataset of 101 human actions classes from videos in the wild," *Center Res. Comput. Vis.*, Univ. Central Florida, Orlando, FL, USA, Tech. Rep. CRCV-TR12-01, Nov. 2012.

[28] Y. Jia *et al.*, "Caffe: Convolutional architecture for fast feature embedding," in *Proc. ACM Int. Conf. Multimedia*, 2014, pp. 675–678.

[29] H. Wang, A. Kläser, C. Schmid, and C. Liu, "Dense trajectories and motion boundary descriptors for action recognition," *Int. J. Comput. Vis.*, vol. 103, no. 1, pp. 60–79, 2013.